\documentclass[12pt]{article}

\usepackage[utf8]{inputenc}
\usepackage{amsmath}
\usepackage{graphicx}
\usepackage{url}
\usepackage[a4paper, total={6in, 8in}]{geometry}
\usepackage{booktabs}
\usepackage{hyperref}

\title{Evaluating Large Language Models with Human Feedback: Establishing a Swedish Benchmark}
\author{Birger Moëll \\ KTH}
\date{\today}

\begin{document}

\maketitle

\begin{abstract}
In the rapidly evolving field of artificial intelligence, large language models (LLMs) have demonstrated significant capabilities across numerous applications. However, the performance of these models in languages with fewer resources, such as Swedish, remains under-explored. This study introduces a comprehensive human benchmark to assess the efficacy of prominent LLMs in understanding and generating Swedish language texts using forced choice ranking. We employ a modified version of the ChatbotArena benchmark, incorporating human feedback to evaluate twelve different models, including GPT-4, GPT-3.5, various Claude and Llama models, and bespoke models like Dolphin-2.9-llama3b-8b-flashback and BeagleCatMunin. These models were chosen based on  their performance on LMSYS chatbot arena and the Scandeval benchmarks. We release the chatbotarena.se benchmark as a tool to improve our understanding of language model performance in Swedish with the hopes that it will be widely used. We aim to create a leaderboard once sufficient data has been collected and analysed.
\end{abstract}

\section{Introduction}
Large language models (LLMs) have demonstrated exceptional capabilities across various applications, significantly advancing the field of natural language processing. However, their effectiveness in low-resource languages remains underexplored. The majority of these models are optimized for English or other high-resource languages, leading to a notable performance disparity when applied to less commonly used languages. This gap in model performance has significant implications:

\begin{itemize}
    \item \textbf{Accessibility:} Individuals who are native speakers of low-resource languages but not proficient in English are less able to benefit from the advancements in AI and machine learning. This creates a barrier to accessing technology that could otherwise support education, business, and communication.
    \item \textbf{Research Limitations:} The lack of scientific investigation into LLM performance across diverse linguistic landscapes means that improvements and innovations are often not tailored to the needs of all potential users. This oversight can perpetuate inequities in technology access and effectiveness.
    \item \textbf{Potential model improvements:} Since models are not primarily tailored for the language, improved understanding in combination with fine-tuning techniques has the potential to significantly advance model performance for the target language.
\end{itemize}

\subsection{Motivation for creating Swedish Chatbot Arena}
Benchmarks for Swedish exist with the excellent Scandeval \cite{nielsen-2023-scandeval} being the primary tool used. ScandEval gives a good indication of model performance but the reliance on automatic benchmarking has some drawbacks, training on the benchmark can skew results and good performing models might not be what is preferred by humans. To adress these issues we introduce the Swedish Chatbot Arena, a Swedish human preference benchmark, where the goal is to assess what models humans prefer. Human can be seen as the gold standard for evaluating LLMs with \cite{chiang2024chatbot} LMSYS chatbot arena being the defacto standard for rating models reliably. Our benchmark draws heavily on the LMSYS benchmark and we have \href{https://github.com/BirgerMoell/SwedishLLMBenchmark}{adopted the source code}  to work with our Swedish model. 

\subsection{Evaluation strategies}
\subsubsection{Subjective evaluation / Human feedback ranking}

Human feedback plays a crucial role in the democratic evaluation of large language models (LLMs). As these models increasingly influence various sectors such as media, education, and even governance, it is essential to ensure that they operate in a manner that aligns with the values and needs of diverse user groups. By involving humans in the voting process on the quality of LLMs, we achieve several democratic objectives:

\begin{itemize}
    \item \textbf{Representation:} Including a broad spectrum of individuals in the evaluation process ensures that the models serve the needs of a diverse population rather than a select few. This inclusivity helps prevent biases that may otherwise emerge if only a limited demographic is considered.
    \item \textbf{Accountability:} Human feedback mechanisms make developers and stakeholders accountable to the end-users. By actively soliciting feedback, model developers are encouraged to make improvements that are transparent and in the best interest of the public.
    \item \textbf{Adaptability:} Through continuous human engagement, LLMs can be dynamically adjusted to meet changing societal norms and expectations. This adaptability is crucial in maintaining the relevance and appropriateness of automated systems in a democratic society.
    \item \textbf{Trust:} When end-users contribute to the evaluation and enhancement of LLMs, it builds trust in the technology. Trust is essential for the wider adoption and acceptance of AI technologies in society.
    \item \textbf{Ethical considerations:} Human judgment is indispensable in navigating the complex ethical landscapes that AI systems, like LLMs, often encounter. Voting on model quality allows for ethical deliberations that purely data-driven approaches may overlook.
\end{itemize}

Swedish Chatbot Arena uses forced ranking between two models to create a benchmark of model performance using ELO ratings. This ensures a fair evaluations of models while ensuring that a broad aspect of the public can be a part in democratically evaluating LLM performance.

\subsection{Evaluation strategies}
\subsubsection{Objective evaluation / Automatic evaluation tools}

Automatic evaluation tools such as Scandeval \cite{nielsen-2023-scandeval} play an indispensable role in the development and refinement of large language models (LLMs). These tools provide a systematic and standardized approach to assessing model performance across multiple benchmarks, offering several distinct advantages:

\begin{itemize}
    \item \textbf{Efficiency:} Automatic tools can evaluate models much more quickly than human-based assessments. This speed is crucial for iterating over model designs, allowing developers to refine and test models continuously without significant delays.
    \item \textbf{Consistency:} These tools apply the same standards and methodologies across different models, ensuring that the evaluations are consistent. This uniformity is essential for fair comparisons between models, facilitating a clearer understanding of each model's strengths and weaknesses.
    \item \textbf{Scalability:} Automated tools can handle large volumes of evaluation tasks simultaneously, making them scalable to the needs of rapidly developing AI technologies. This capability is particularly valuable in the context of large-scale LLMs, which require extensive testing across diverse datasets.
    \item \textbf{Repeatability:} Automatic evaluations can be repeated under the same conditions to verify results. This repeatability helps in ensuring the reliability of the evaluations and allows researchers to systematically track improvements over time.
    \item \textbf{Benchmarking:} Objective evaluation provide benchmarks that are specifically designed to test various aspects of LLM performance, such as linguistic understanding, contextual awareness, and the ability to generate coherent text. This targeted assessment helps pinpoint specific areas where a model may need further development.
\end{itemize}

The use of automated evaluation tools is crucial in the LLM development lifecycle as they provide rapid feedback that is essential for timely and effective model training and refinement. By leveraging these tools, developers can accelerate the development process and enhance the overall quality and effectiveness of their models. 

\subsection{Objective and subjective evaluation}
We believe that combining objective evaluation with a tool such as Scandeval with subjective evaluation with a tool such as Swedish Chatbot Arena is the most efficient way to get an accurate assessment of model performance.

\section{Method}

\subsection{Evaluation platform}
Our software evaluation platform is a fork of the LMSYS project which currently runs on a RTX-4090 GPU with enough memory to serve two open source models.

\subsection{Model Selection}
Twelve language models were initially selected for inclusion based on evaluation on LMSYS and Scandeval on Swedish-language tasks. Limitations such as availability in the EU / availability through an API limited model selection. A majority of models were served through an API, 10 models while two models Dolphin-2.9-llama3b-8b-flashback and BeagleCatMunin was served through a local GPU.

The models currently included in the benchmark can be viewed in table 1.

\begin{table}[h]
\centering
\caption{Language Models Selected}
\begin{tabular}{lccccr}
\toprule
Model & Model Size & Open License & LMSYS & Scandeval & Training \\
\midrule
GPT-4o & - & No & 1 & - & - \\
GPT-4-5 & - & No & 2 & - & - \\
GPT-4 & - & No & 2 & 1.07 & - \\
GPT-3.5 & - & No & 29 & 1.88 & - \\

Claude Opus & - & No & 2 & - & - \\
Claude Sonnet & - & No & 8 & - & - \\
Claude Haiku & - & No & 13 & - & - \\

Llama70b-Instruct & 70b & Yes & 7 & 1.82 & base + instruction \\
Llama-8b-Instruct & 8b & Yes & 17 & 2.52 & base + instruction \\

AI-Sweden-20b & 20b & No & - & 3.28 & base + fine-tune \\
Dolphin-flashback* & 8b & Yes & - & 2.36 & fine-tune + merge \\
BeagleCatMunin & 8b & Yes & - & 2.37 &  fine-tune + merge \\

\bottomrule
\end{tabular}
*Full name, Dolphin-2.9-llama3b-8b-flashback

For Scandeval, lower rank is better.
\end{table}

\subsection{Open AI models}
The models GPT-3.5 and GPT-4, GPT-4.5 and GPT-4o was natural to include since they are the most well used language models with GPT-3.5. powering ChatGPT and GPT-4 being the most capable model in most evaluations and GPT-4o ranking number one on the LMSYS Chatbot Arena  \cite{chiang2024chatbot}.

\subsection{Anthropic Models}
Anthropic released Anthropic version 3 on March 4th, 2024 with the models Haiki, Sonnet and Opus.
Training techniques includes pretraining on large data to acquire language capabilities through word prediction, as well
as human feedback techniques. \cite{anthropic2023claude} Anthropic used a technique
called Constitutional AI that uses RLAIF for feedback for alignment. \cite{bai2022constitutional}. 

The Opus overtook GPT-4 to become the best performing model on the LMSYS benchmark before the release of GPT-4.5. The models are highly capable and although they are not reviewed on Scandeval benchmark, they deserve to be reviewed further in this evaluation.

\subsection{Llama Models}
On the 18th of April 2024 Meta open sourced Llama 3. Llama 3 was trained on 15 trillion tokens collected from public sources. Over 5\% of the data consists of high quality non English data. \cite{metallama3}
Both the Llama 70b and the Llama 8b models are capable with high scores on both LMSYS chatbot arena and Scandeval. 

\subsubsection{Dolphin-2.9-llama3b-8b-flashback}
Several fine-tunes and merges have been done on Llama3 models for Swedish. Dolphin-2.9-llama3b-8b-flashback is a high performing example of a lllama-3 fine tune. The model is a merge of timpal0l/Llama-3-8B-flashback-v1 and cognitivecomputations/dolphin-2.9-llama3-8b which in turn are fine-tunes on Swedish language data (Flashback) and English language data. Notably the underlying dolphin model is uncensored to remove alignment and bias. This model is part of the the benchmark for research purpose but can generate unethical responses.

\subsection{Open models}

\subsubsection{BeagleCatMunin}
BeagleCatMunin is similar to Dolphin-2.9-llama3b-8b-flashback by being a merge of fine-tuned models. The model is based on the Mistral-7b architecture. The underlying Swedish model has been trained on flashback data and it is merged with a highly performant Scandinvian model, RJuro/munin-neuralbeagle-7b.

\subsubsection{AI-Sweden-20B}
AI-Sweden 20b is a model created by AI Sweden and trained on the Nordic Pile dataset.

\section{Types of models evaluated}
Since the benchmark contains both open source and closed source, not all knowledge regarding training of the models is available. However, based on information regarding the open models some conclusions regarding models can be drawn.

\subsection{Instruction tuned models}
Instruction tuning involves creating a dataset of instructions (prompts) and answer for a base trained model to follow. Instruction tuning has been shown to increase model performance and Llama 8b Instruct and Llama 70b Instruct are examples of instruction tuned models in this benchmark.

\subsection{Reinforcement Learning from Human Feedback}
Reinforcement Learning from Human Feedback is a step in training where a model uses human feedback to improve model performance. Most models included in this benchmark has at some point been trained with RLHF. Notably Dolphin-2.9-llama3b-8b-flashback is the only model where an attempt has been made to remove preferences made from RLFH training.

\subsection{Merged models}
Merging\cite{yadav2023tiesmerging} is a technique where models are combined to form a new model by combining the weighs inside the neural network directly. This data free training method has several benefits with perhaps the biggest one being the ability to merge a model without the need for a GPU. This in combination with the high quality performance of merge models have made them a common technique for working with models. 


\section{Next steps}
We hope that everyone will help out in our work on improving Swedish Language models by evaluating models on chatbotarena.se, the home of our benchmark.

Once evaluations are done and we have gathered enough data, we will present a benchmark of the most performant models for Swedish.

\subsection{Data}
Our benchmark relies on forced binary ranking of two models which gives as output a preference for an answer over another. This output is useful both as assessment of model performance but also as a way to create Human Preference Data that can be used to improve model performance.

\subsection{How can you help?}
The goal of our project is to improve access to high quality language models in Swedish. The easiest way to help is to simply evaluate models. We also welcome pull requests and feedback on our repo\footnote{\href{https://github.com/BirgerMoell/SwedishLLMBenchmark}{https://github.com/BirgerMoell/SwedishLLMBenchmark}}.

\bibliographystyle{plainnat} 
\bibliography{references} 

\begin{thebibliography}{6}
\providecommand{\natexlab}[1]{#1}
\providecommand{\url}[1]{\texttt{#1}}
\expandafter\ifx\csname urlstyle\endcsname\relax
  \providecommand{\doi}[1]{doi: #1}\else
  \providecommand{\doi}{doi: \begingroup \urlstyle{rm}\Url}\fi

\bibitem[{Anthropic}(2023)]{anthropic2023claude}
{Anthropic}.
\newblock Model card for claude 3.
\newblock
  \url{https://www-cdn.anthropic.com/de8ba9b01c9ab7cbabf5c33b80b7bbc618857627/Model_Card_Claude_3.pdf},
  2023.
\newblock Accessed: 2024-05-02.

\bibitem[Bai et~al.(2022)Bai, Kadavath, Kundu, Askell, Kernion, Jones, Chen,
  Goldie, Mirhoseini, McKinnon, Chen, Olsson, Olah, Hernandez, Drain, Ganguli,
  Li, Tran-Johnson, Perez, Kerr, Mueller, Ladish, Landau, Ndousse, Lukosuite,
  Lovitt, Sellitto, Elhage, Schiefer, Mercado, DasSarma, Lasenby, Larson,
  Ringer, Johnston, Kravec, Showk, Fort, Lanham, Telleen-Lawton, Conerly,
  Henighan, Hume, Bowman, Hatfield-Dodds, Mann, Amodei, Joseph, McCandlish,
  Brown, and Kaplan]{bai2022constitutional}
Yuntao Bai, Saurav Kadavath, Sandipan Kundu, Amanda Askell, Jackson Kernion,
  Andy Jones, Anna Chen, Anna Goldie, Azalia Mirhoseini, Cameron McKinnon,
  Carol Chen, Catherine Olsson, Christopher Olah, Danny Hernandez, Dawn Drain,
  Deep Ganguli, Dustin Li, Eli Tran-Johnson, Ethan Perez, Jamie Kerr, Jared
  Mueller, Jeffrey Ladish, Joshua Landau, Kamal Ndousse, Kamile Lukosuite,
  Liane Lovitt, Michael Sellitto, Nelson Elhage, Nicholas Schiefer, Noemi
  Mercado, Nova DasSarma, Robert Lasenby, Robin Larson, Sam Ringer, Scott
  Johnston, Shauna Kravec, Sheer~El Showk, Stanislav Fort, Tamera Lanham,
  Timothy Telleen-Lawton, Tom Conerly, Tom Henighan, Tristan Hume, Samuel~R.
  Bowman, Zac Hatfield-Dodds, Ben Mann, Dario Amodei, Nicholas Joseph, Sam
  McCandlish, Tom Brown, and Jared Kaplan.
\newblock Constitutional ai: Harmlessness from ai feedback, 2022.

\bibitem[Chiang et~al.(2024)Chiang, Zheng, Sheng, Angelopoulos, Li, Li, Zhang,
  Zhu, Jordan, Gonzalez, and Stoica]{chiang2024chatbot}
Wei-Lin Chiang, Lianmin Zheng, Ying Sheng, Anastasios~Nikolas Angelopoulos,
  Tianle Li, Dacheng Li, Hao Zhang, Banghua Zhu, Michael Jordan, Joseph~E.
  Gonzalez, and Ion Stoica.
\newblock Chatbot arena: An open platform for evaluating llms by human
  preference, 2024.

\bibitem[{Meta}(2023)]{metallama3}
{Meta}.
\newblock Llama 3 release blog.
\newblock \url{https://ai.meta.com/blog/meta-llama-3/}, 2023.
\newblock Accessed: 2024-05-02.

\bibitem[Nielsen(2023)]{nielsen-2023-scandeval}
Dan Nielsen.
\newblock {S}cand{E}val: A benchmark for {S}candinavian natural language
  processing.
\newblock In Tanel Alum{\"a}e and Mark Fishel, editors, \emph{Proceedings of
  the 24th Nordic Conference on Computational Linguistics (NoDaLiDa)}, pages
  185--201, T{\'o}rshavn, Faroe Islands, May 2023. University of Tartu Library.
\newblock URL \url{https://aclanthology.org/2023.nodalida-1.20}.

\bibitem[Yadav et~al.(2023)Yadav, Tam, Choshen, Raffel, and
  Bansal]{yadav2023tiesmerging}
Prateek Yadav, Derek Tam, Leshem Choshen, Colin Raffel, and Mohit Bansal.
\newblock Ties-merging: Resolving interference when merging models, 2023.

\end{thebibliography}

\end{document}